\documentclass[12 pt]{article}
\usepackage{graphicx}
\pdfoutput=1
\usepackage{amsmath}
\usepackage[ruled,vlined]{algorithm2e}

\begin{document}
	
\title{Modeling Systems with Machine Learning based Differential Equations}
\author{	P. Garc\'{\i}a\\
\small		Laboratorio de Sistemas Complejos,\\
\small		Departamento de F\'{\i}sica Aplicada,\\
\small		Facultad de Ingenier\'{\i}a, 
\small		Universidad Central de Venezuela.\\
		
\small		and\\
		
\small		Red Iberoamericana de Investigadores en
\small		Matemáticas Aplicadas a  Datos. AUIP.
}

\date{}
	
\maketitle
\thispagestyle{empty}

\begin{abstract}
	The prediction of behavior in dynamical systems, is frequently subject to the design of models. 
	When a time series obtained from observing the system is available, the task 
	can be performed by designing the model from these observations without additional assumptions or by assuming a preconceived structure in the model, with the help of additional information about the system. In the second case, it is a question of adequately combining theory with observations and subsequently optimizing the mixture.
	
	In this work, we proposes the design of time-continuous models of dynamical systems as solutions of differential equations, from non-uniform sampled or noisy observations, using machine learning techniques.
	
	The performance of strategy is shown with both, several simulated data sets and experimental data from Hare-Lynx population and Coronavirus 2019 outbreack.
	
	Our results suggest that this approach to the modeling systems, can be an useful technique in the case of synthetic or experimental data.
\end{abstract}

\section{Introduction}
It's a known fact, that research at intersection area between two scientific disciplines, can produces remarkable results for both. Physics and Machine Learning are an example of this. 
The use of machine learning techniques in the characterization of systems or in the prediction of their behavior has been a constant in recent years \cite{Buchanan, Carleo}.
In return, physics has collaborated with the explaining, from statistical mechanics point of view, of the training mechanism of neural networks \cite{Bahri}, among many others contributions.

In the particular case of subareas of such as Dynamical Systems and Neural Networks, there are developments that are already notable and although there are many strategies that can be inscribed in this intersection, mixed strategies between differential equations theory and neural networks, which suggest that there exists a duality between differential equations and many deep neural networks that is worth trying to understand.

With the idea of establishing some order in this type of results, we will divide them into two classes:
(i) the estrategies that use neural networks to solve differential equations \cite{Lagaris, Sirignano, Raissi} and (ii) the methodologies that tells us how modern differential equation solvers can simplify the architecture of many neural network \cite{He, Chen}.

Particularly, in the work of Chen et. al. they define a hybrid of neural net and differential equation, the {\it Neural Ordinary Differential Equation} (Neural ODEs) with striking properties including excellent accuracy and greatly reduced memory requirements.
On the seccond one of those classes, we will focus our results and comparisons.

This association between machine learning and differential equations is very likely to be beneficial for physics and other scientific disciplines, since in our times from the 17th century, differential equations, or its discreet versions, are the most popular way of representing dynamical systems. Particularly in physics, the label of "most popular" belongs to the initial value problems. From this we start our purpose.

Given a initial value problem for a general ordinary differential equation:

\begin{eqnarray}
\frac{dx(t)}{dt} &=& f(x(t),t,\theta),  \\
x(t_0) &=& x_0 \nonumber
\label{ODE}
\end{eqnarray}

\noindent
where $ \theta $ is a vector of parameters.
A problem that appears frequently, in many areas, when systems are represented in this way, consists of: given $ x (t_0) $, calculate $ x (t_1) $, i.e. an initial value problem, whose formal solution is:

\begin{equation}
x(t) = x(t_0) + \int_{t_0}^{t} f(x(t),t,\theta) dt
\label{ODE-sol}
\end{equation}

In the case in which $f$ cannot be integrated analytically, an approximation to the solution consists of numerically solving the previous integral. In general, regardless of the numerical method  used, numerical solution of (\ref{ODE-sol}) can be written as:
\begin{equation}
x(t) = ODESolve\left( f(x(t),t,\theta),x_0, t_0, t, \theta \right)
\end{equation}			

Among all these works we will focus, to motivate ours, on the results of He \cite{He} and Chen \cite{Chen}, where it is shown that can be designed a neural network that accounts for the rate of change of the system when it goes from state $x(t_{n})$ to state $x(t_{n+1})$. The neural network architecture that results from this idea is known as ResNet and is the prehistory of the work \cite{Chen}, where the authors introduce a new family of deep neural network models that instead of specifying a discrete sequence of hidden layers, parameterize the derivative of the hidden state using a neural network.

Both approaches involve the approximation of the integral in (\ref{ODE-sol}) with a particular integration scheme, within some neural network training scheme. 	

With this in mind we propose to design a continuous model from discrete observations, using strategies from machine learning regression (different from Neural Networks) to provides an efficient way to use noisy, non-uniformly sampled data to determine a reliable and continuous time model.
Non-uniformly sampled data can appear when sampling at the Nyquist frequency is not efficient \cite{Craven}, when there are incompleteness due to difficulties in data collection \cite{Fournet} and continuous models over time are useful in system identification \cite{Bekiroglu}, forecast \cite{Fournet} and control \cite{} of such systems.

Continuous-time system identification from non-uniform sampled data has
also been considered using B-splines functions \cite{Gillberg}, Kalman filters \cite{Ding},  Box-Jenkins  \cite{Chen-Garnier} and expectation-maximization algorithm \cite{Ding}.

In our case, we have used a kernel-besed regression technique to approximate  (\ref{ODE})and in order to show how we develop the before mentioned idea, this paper was organized as follow: in section 2 we setting the problem of nonlinear modeling, in section 3 we sketch the solution, in section 4 we show examples of the performance of the proposed  method. Finally in section 5 we summarize our observations and give conclusions.		

\section{Setting the problem}
We set our problem as the design of a Machine Learning regression version of $f$ in (\ref{ODE}) 

\begin{eqnarray}
\frac{dx(t)}{dt} &=& \tilde f(x(t),\omega), \\
x(t_0) &=& x_0  \nonumber
\label{ODEapp}
\end{eqnarray}	

\noindent
from a time series $\{x(t_n)\}_{n=1}^N$ with $x \in \Re^n$, non-uniformly sampled, from the observation of the unknown system (\ref{ODE}), to construct a model, 
such that, finite segments of the trajectories of (\ref{ODE}) are kept as close as possible to segments of the same length, of trajectories of (\ref{ODEapp}).

Once we have chosen the space of functions, of which $\tilde f$ is an element, and defined a norm in that space, it is possible\cite{Garcia} to set a optimization problem:

\begin{equation}
\omega = \operatorname*{argmin}_\omega {\left( L(e,\omega) \right) },
\label{Optimization}
\end{equation}

\noindent
from the functional of error: 

\begin{equation}
L(e,\omega) = \frac{1}{2} \left( \sum_{i=1}^{N-1} e_i^2 + \lambda \parallel \tilde f(x(t),\omega) \parallel \right) ,
\label{Lagrangian}
\end{equation}

\noindent
where, $e_i = x(t_{i+1})- x(t_{i}) + \delta t_i~ \tilde f(x(t_{i}),\omega)$, with $\delta t_i = t_{i+1}-t_i$ ~a variable-length interval of sampling, in contrast with the usual assumption in regression techniques that suppose the signal or time series is uniformly sampled in time. 

\section{Kernel regression approach to ODEs}
The error representation (\ref{Lagrangian}) assumes a discrete version of (\ref{ODEapp}) (using the Euler scheme) that we had written as:
\begin{eqnarray}
x_{i+1} &=& x_i + \delta t_i  ~\tilde f(x_i, \omega)  \nonumber
\end{eqnarray}	

\noindent
where $x(t_{i}) = x_i$.
As an alternative way, to the neural networks proposed in \cite{He, Chen}, to approximate $\tilde f$, in this work a kernel-based regression scheme known as {\it Kernel Ridge Regression} \cite{Scholkopf}  is used. This scheme, in addition to being conceptually simple, offers a low computational cost solution strategy for (\ref{Optimization}). 

The proposed kernel method \cite{Scholkopf} belongs to a class of machine learning algorithms implemented for many different inferential tasks and application areas. The inference models, trained using the kernel approach, allow to obtain new data representations for the training patterns, by embedding such observations, into a new feature space, where, using techniques from optimization, mathematical analysis and statistics, one can extrapolate interesting properties, required to the inference for new data.

Without losing generality, it can be assumed that $\tilde f$ belongs to a Hilbert space $\mathcal H$, so that it can be written as:

\begin{equation}
\tilde f(x_i) = \sum_{n=0}^{\infty} \alpha_n \varphi_n(x_i),
\label{representation}
\end{equation}

\noindent
where $ {\varphi_n} $ is a base for $ \mathcal H $ and $ {\alpha_n} $ is the set of coefficients to fit.

Given that the cost functional (\ref{Lagrangian}) and the approximation (\ref {representation}), the quality of our approximation will look like:

\begin{equation}
L(e,\omega) = \frac{1}{2}\left( \sum_{i=0}^{N} \left[ x_{i+1} - x_{i}  + \delta t_i \tilde f(x_i, \omega)
\right]^2 + \lambda  \sum_{n=0}^{\infty} \alpha^2_n \right)
\end{equation}

\noindent
for orthonormal basis.

If we want to determine the set $ \{\alpha_n \} $ such that this functional is minimum, one way is to satisfy $ \frac{\partial L} {\partial \alpha_j} = 0 $, so :

\begin{eqnarray}
\sum_{i=0}^{N} \left[ x_{i+1} -  x_{i} + \delta t_i \tilde f(x_i, \omega)\right]
\frac{\partial \tilde f}{\partial \alpha_j} + \lambda \alpha_j &=& 0 \nonumber \\
\sum_{i=0}^{N} \left[ \delta x_{i} + \delta t_i ~\tilde f(x_{i}, \omega)\right]
~ \varphi_j(x_i)+ \lambda \alpha_j &=& 0 \\ \nonumber
\sum_{i=0}^{N} \left[ \delta x_{i} + \delta t_i ~\sum_{n=0}^{\infty} \alpha_n \varphi_n(x_i)\right]
~ \varphi_j(x_i) + \lambda \alpha_j &=& 0 
\end{eqnarray}

\noindent
where $ \delta x_{i} = x_{i + 1} - x_{i} $. From the orthogonality of the set $\{\varphi_n \}$ we have:

\begin{equation}
\sum_{i=0}^{N}  \delta x_{i} ~ \varphi_j(x_i) + \beta_j~ \alpha_j 
+ \lambda \alpha_j = 0
\end{equation}

\noindent
where

\begin{equation}
\beta_j =  \sum_{i=0}^{N}  \delta t_{i} ~ \varphi_j(x_i) ~ \varphi_j(x_i) 
\end{equation}

\noindent
or

\begin{equation}
\sum_{i=0}^{N}  \delta x_{i} ~ \varphi_j(x_i)  +  \beta_j \alpha_j 
+ \lambda \alpha_j = 0
\end{equation}

So that:

\begin{equation}
\alpha_j = \frac{1}{(\beta_j + \lambda)} \sum_{i=0}^{N} \delta x_{i} ~\varphi_j(x_i)
\end{equation}

Thus, our approximation can be written as:

\begin{equation}
\tilde f(x) = \sum_{n=0}^{\infty} \left [ \sum_{i=0}^{N} \frac{1}{(\beta_n + \lambda)} \delta x_{i}~ \varphi_n(x_i)  \right ] \varphi_n(x), 
\end{equation}

\noindent
or alternatively: 

\begin{equation}
\tilde f(x) =   \sum_{i=0}^{N} \left [ \sum_{n=0}^{\infty} \frac{1}{(\beta_n + \lambda)} \varphi_n(x_i) \varphi_n(x) \right ]   \delta x_{i} , 
\end{equation}

Let's call:

\begin{equation}
K(x_i,x) =   \sum_{n=0}^{\infty}  \frac{1}{(\beta_n + \lambda)}  \varphi_n(x_i) \varphi_n(x), 
\end{equation}

\noindent
so that $ \tilde f $ could be represented as:

\begin{equation}
\tilde f(x) =  \sum_{i=0}^{N} \omega_i K(x_i,x),
\end{equation}

\noindent
where $ \omega_i = \delta x_{i}$ and $ K $ is a reproductive kernel of the space $ \mathcal H $. If we have infinite data, then the sum (\ ref {}) can be carried out. If not, it is possible to use a regression scheme that allows estimating the $ w_i $, using a particular kernel.

It is worth noting here, that the non-linear approximation problem of $ \tilde f $ is transformed into a linear, i. e., the determination of $ \omega_i $, using only elements ($ x_i $) of the original space.

Although there are many more alternatives,  here we chose $K$ as the Gaussian kernel,

\begin{equation}
K(x_i, x_j) = \exp \left(-\frac{(x_i-x_j)^2}{2 s^2}\right),
\end{equation}	

\begin{figure}[h]
	\begin{center}
		\includegraphics[width=10.5cm, height=7.5cm]{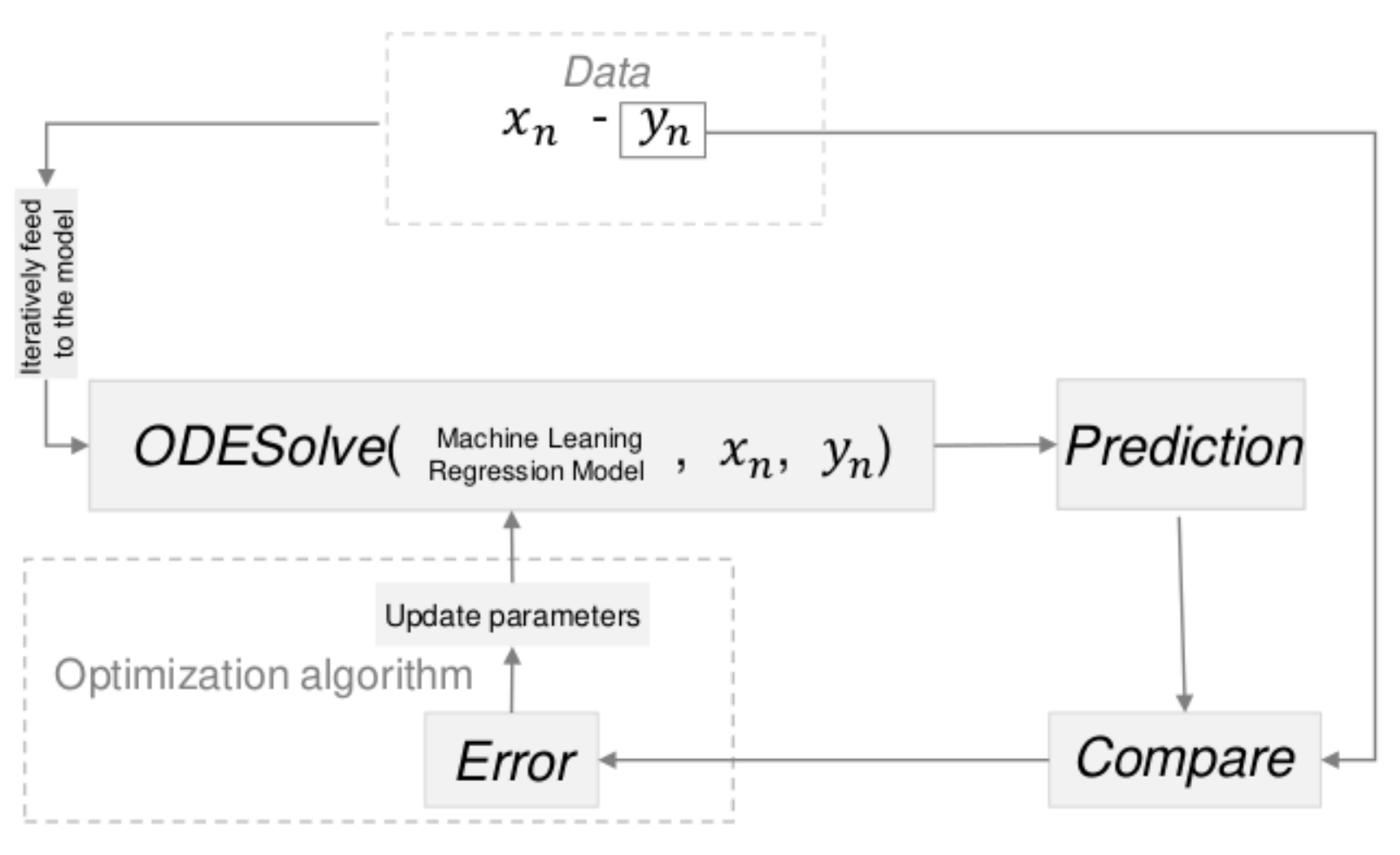}
	\end{center}
	\caption{A graphical representation of the idea behind the modeling scheme.}
	\label{Fig1}
\end{figure}

Once $ K $ is chosen, the nonlinear regression problem of $ \tilde f $ becomes a linear regression problem, i. e., the determination of $ \omega_i $. Althoug this last task can be done using several online algorithms, here we will use Ridge regression. This scheme, is theoretically simple to interpret \cite{Garcia}, and very easy to implement.  The algorithm \ref{algorithm1}, shows the implementation of the strategy shown in Figure \ref{Fig1}.

\begin{algorithm}[h]
	\SetAlgoLined
	Initialize variables: $\omega_n$, $\epsilon$ and $\delta t$.\\
	\While{stopping criterion not met}{
		
		\For{$n = 1$ \KwTo $N$}{
			
			Observe input $x_n$. \\
			Predict output $y_{n+1}$: \\
			$y_{n}=x_n$,\\
			\While{ $\|y_{n}-x_{n+1}\| \leq \epsilon$ } {
				$y_{n+1} \gets y_n + \delta t  ~\sum_{i = 1} ^{N}{\omega_i ~ K(y_n,x_i)}$ 
				\smash{\makebox[0pt][l]{$\qquad
						\left.\begin{array}{c} \strut\\\strut\\\strut\\\strut\\\strut\end{array}\right\} {ODESolver}$$(x_0,t_0,t,\omega)$}
				}\\
				$y_{n}$ $\gets$ $y_{n+1}$ 
			}
			Observe true output $x_{n+1}$.\\
			Update solution $(\omega_n)$ based on $L(e_n)$, with $e_n = x_{n+1}-y_{n+1}$.		
		}	
	}
	\caption{Kernel Ordinary Differential Equation}
	\label{algorithm1}
\end{algorithm}	

This is a very general strategy, with the condition that scheme can be stated as on-line scheme, i. e., where the data elements are presented to the method one each time.

\section{Numerical results}
In order to illustrate how many general are the strategy, we will use several dynamical systems: linear, nonlinear periodic and chaotic, in two and three dimensions. 
In all the previous cases, the training data comes from numerical simulations of the systems. Additionally we present results for two of those cases using real data. In all examples, $M$ samples of the evolution of the system are considered using a sampling rate $ \delta t$ and a random sample of $N$ data is taken from this data. The resulting time serie is finaly standardized.

The {\it capacity of generalization} of the fitted model is qualitatively shown, in all cases, as an orbit, generated from the same initial condition, using the fitted model. In the case of two dimensional systems, we also generate, using the model fitted, a phase portrait. There, it is shown how the flow of many initial conditions converges approximately to the observed orbit.

In the case of simulated data, the results will be displayed in the form of two figures: for uniformly sampled and no-uniformly sampled data.

\subsection{Simulated data}
We show the performance of the method using uniformly and nonuniformly sampled data from: Planar linear system, like in \cite{Chen}, Lotka-Volterra system, Susceptible-Infected-Recovered (SIR) model, and Chua's chaotic system.

\subsubsection{Planar linear system}
A linear system with an asymptotic stable fixed point at $x = 0$. 
\begin{eqnarray}
\dot{x}(t) & = & \alpha ~x(t) + \beta ~y(t)  \nonumber \\
\dot{y}(t) & = &  \gamma ~x(t) + \delta ~y(t)
\end{eqnarray}	

With parameters $\alpha =1$, $v =4$, $\gamma =-2$ and $\delta =2$. Here we use as training data $N=100$ points data obtained by $4$-th order Runge-Kutta method.

\begin{figure}[h]
	\begin{center}
		\includegraphics[width=5cm,height=5cm]{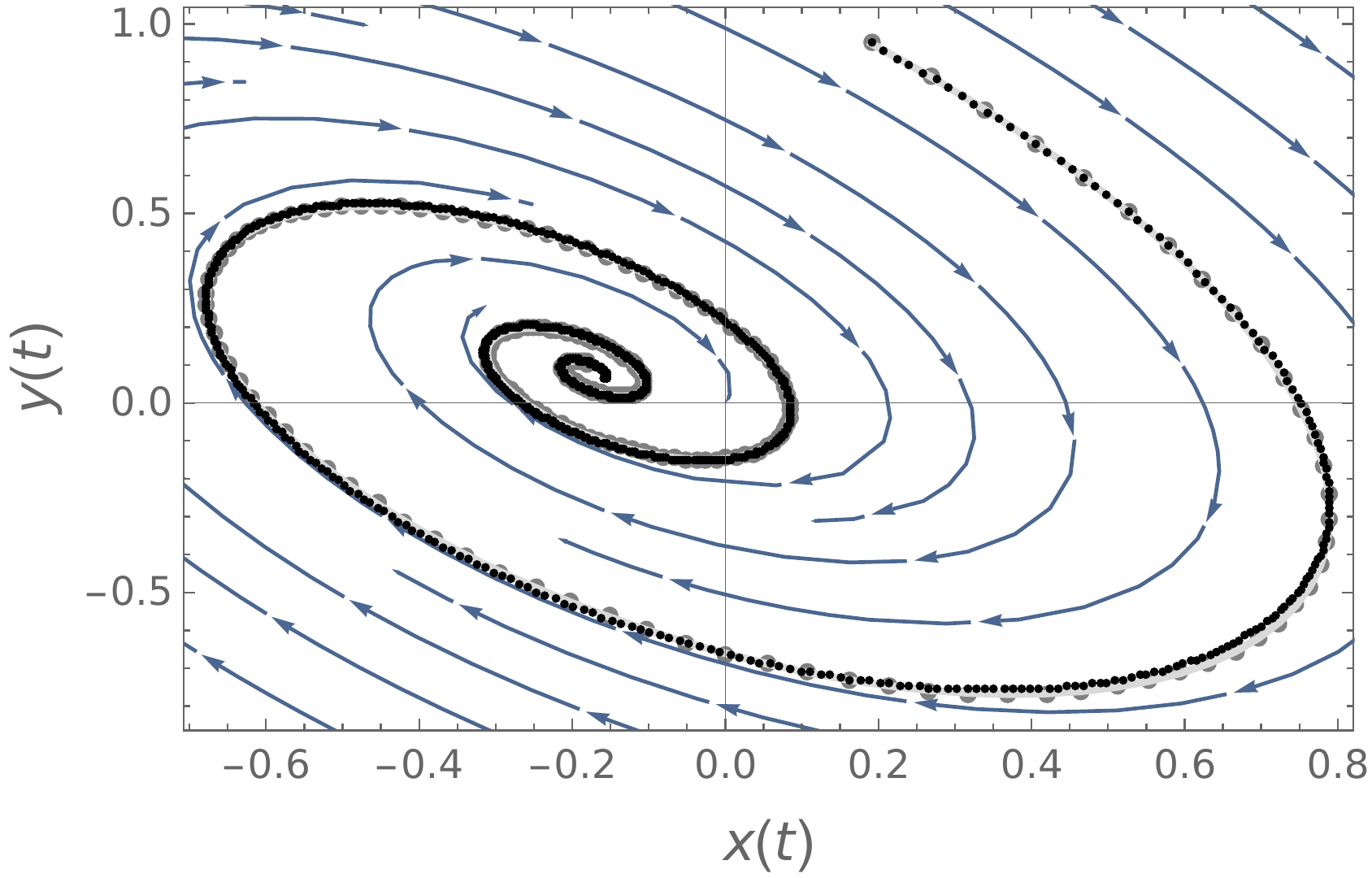}
		\includegraphics[width=5cm,height=5cm]{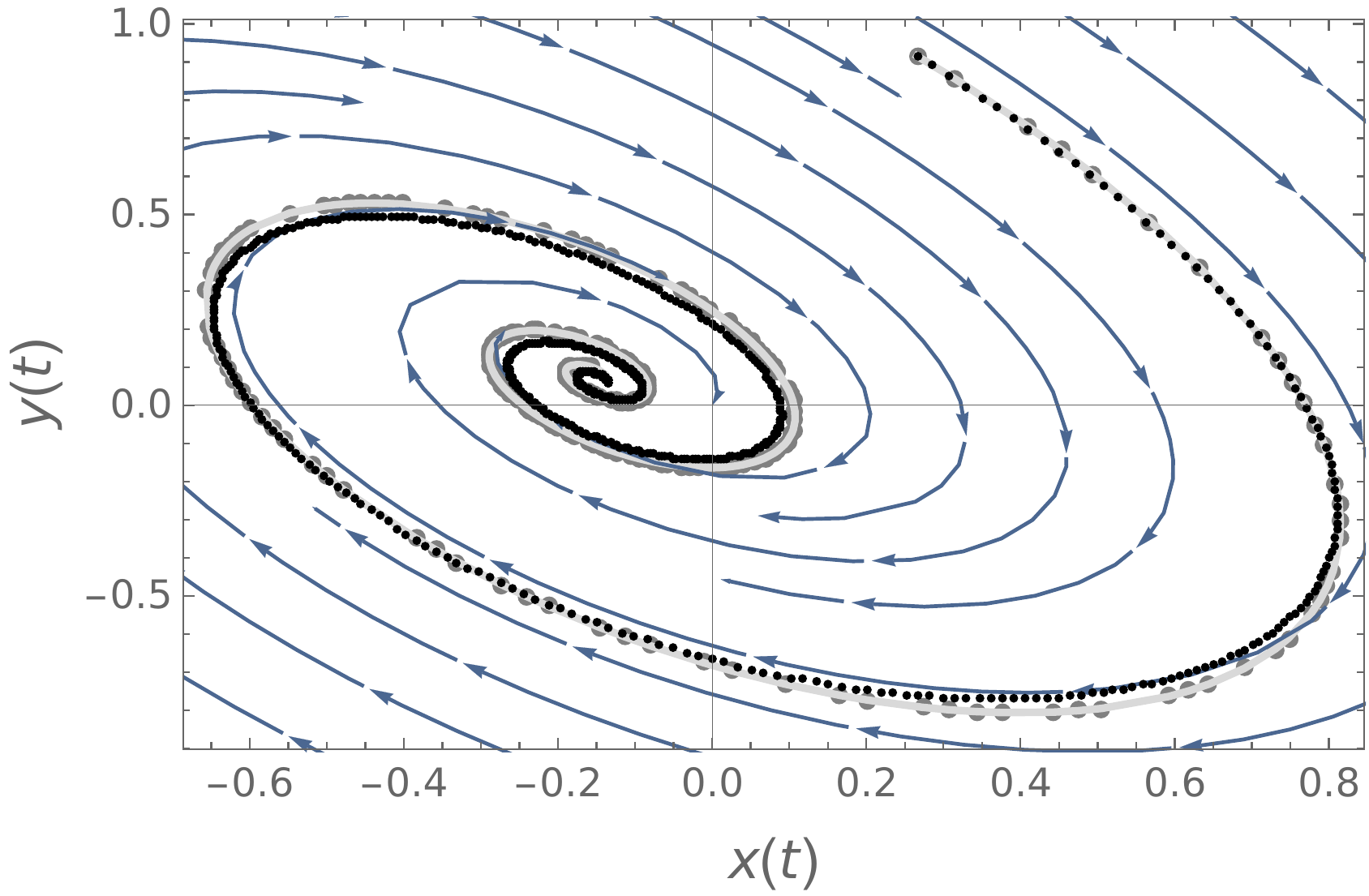} 
		\caption{Planar linear system. The big lightgry points shows the simulates data and the black pints the orbit generated by the model from the first element of the data. The left figure shows the result using uniformly sampled data and the right figure, the same with nonuniformly sampled data.}
	\end{center}
	\label{Fig2}
\end{figure}	

Figure \ref{Fig2} compares the observed and predicted orbit, from the same initial condition and shows the phase protrait generated by the model in the case of this system.

\subsubsection{Lotka-Volterra system}		
The Lotka–Volterra equations, also known as the predator–prey equations, are a pair of first-order nonlinear differential equations, frequently used to describe the dynamics of biological systems in which two species interact, one as a predator and the other as prey. The populations change through time according to the pair of equations:		
\begin{eqnarray}
\dot{x_1}(t) & = &  \alpha  ~x(t)-\beta  ~x(t) y(t)  \nonumber \\
\dot{x_2}(t) & = &  \delta  ~x(t) y(t)-\gamma  ~y(t)
\end{eqnarray}
\noindent
with, $\alpha = 0.1$, $\beta = 0.02$, $\delta = 0.01$ and $\gamma = 0.3$.	Here we use as training data $N=500$ points data obtained by $4$-th order Runge-Kutta method.

\begin{figure}[h]
	\begin{center}
		\includegraphics[width=5cm,  height=5cm]{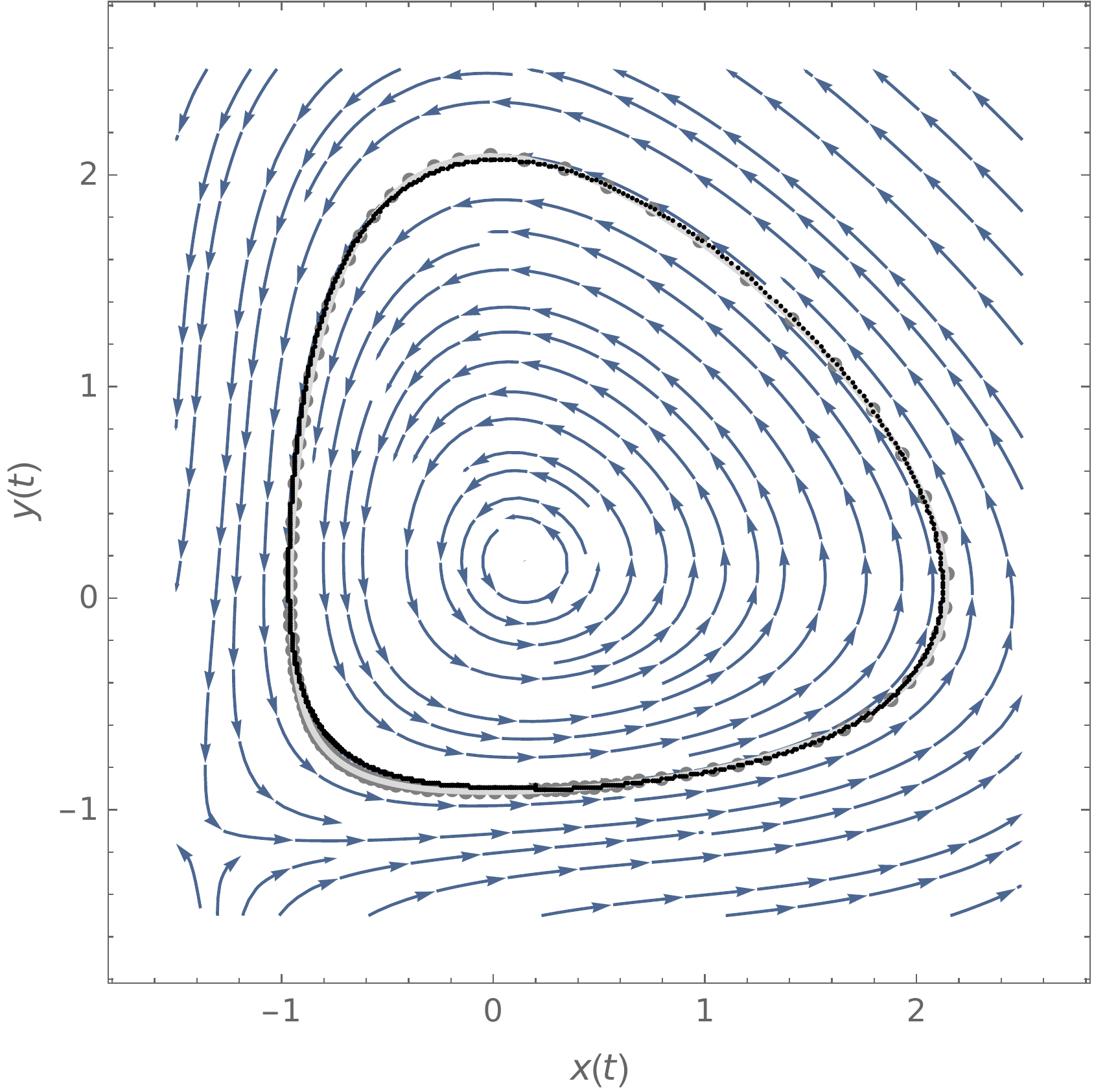}
		\includegraphics[width=5cm, height=5cm]{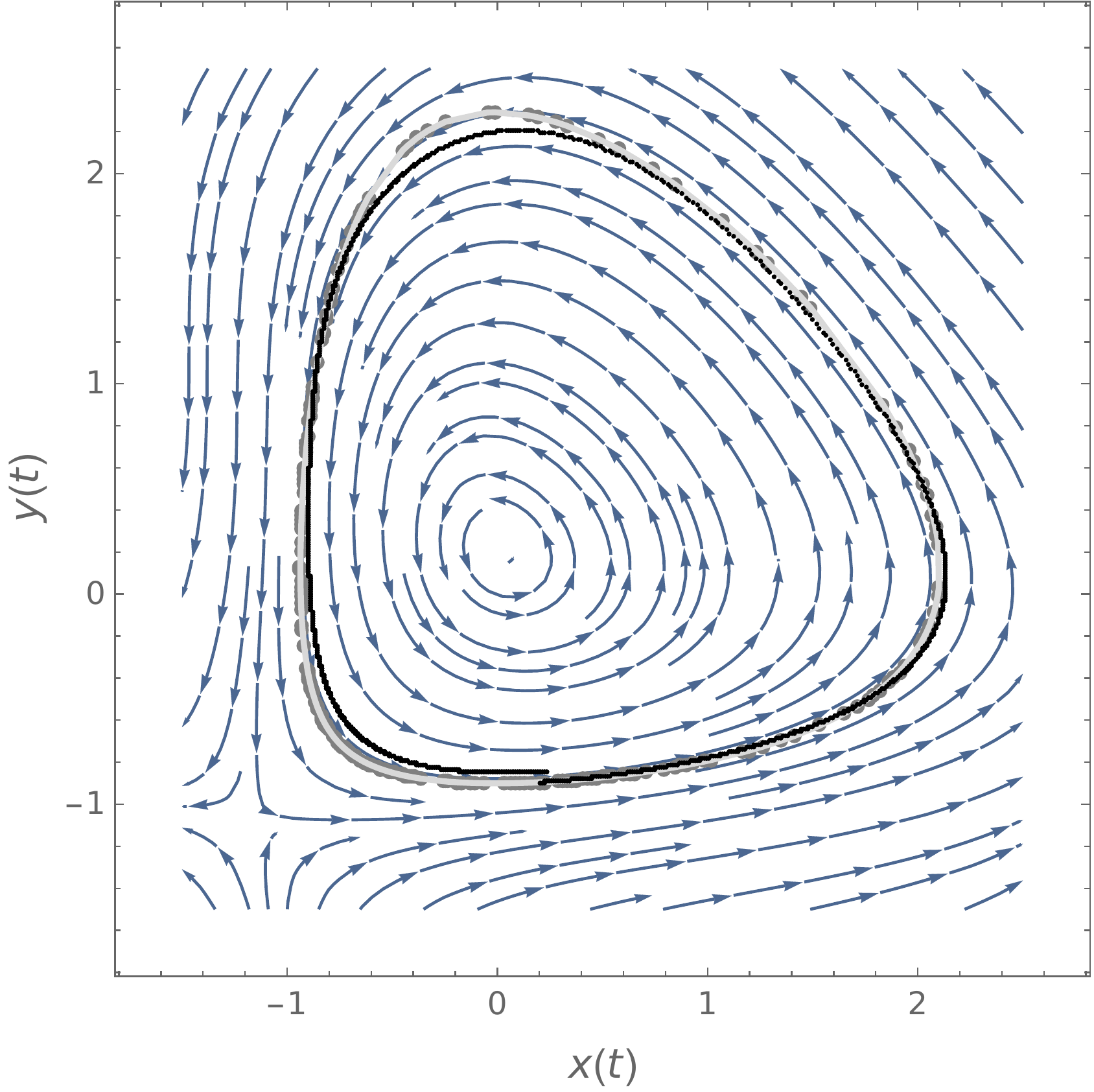}
	\end{center}
	\caption{Lotka-Volterra system. The big lightgry points shows the simulates data and the black pints the orbit generated by the model from the first element of the data. The left figure shows the result using uniformly sampled data and the right figure, the same with nonuniformly sampled data.}
	\label{Fig3}
\end{figure}	

Figure \ref{Fig3}. compares the observed and predicted orbit, from the same initial condition and shows the phase protrait generated by the model in the case of this system

\subsubsection{SIR model}
The SIR model is one of the simplest compartmental models, and many models are derivatives of this basic form. The model consists of three state variable: S the number of susceptible individuals, I the number of infectious individuals and R for the number of removed (and immune) or deceased individuals. 		
It is assumed that the number of deaths is negligible with respect to the total population. This compartment may also be called "recovered" or "resistant". This model is reasonably predictive for infectious diseases that are transmitted from human to human, and where recovery confers lasting resistance, such as measles, mumps and rubella.	

\begin{figure}[h]
	\begin{center}
		\includegraphics[width=5cm, height=5cm]{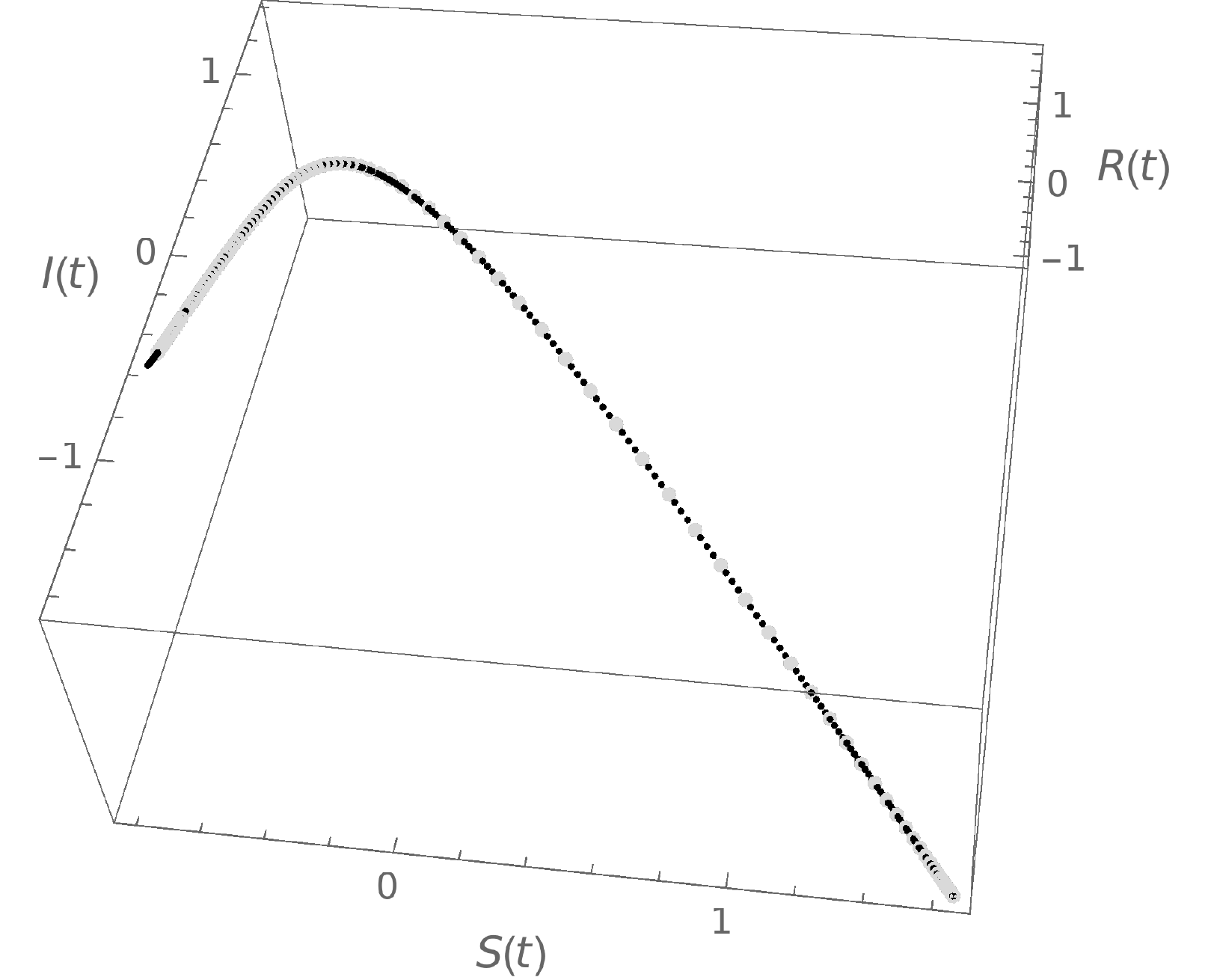}
		\includegraphics[width=5cm, height=5cm]{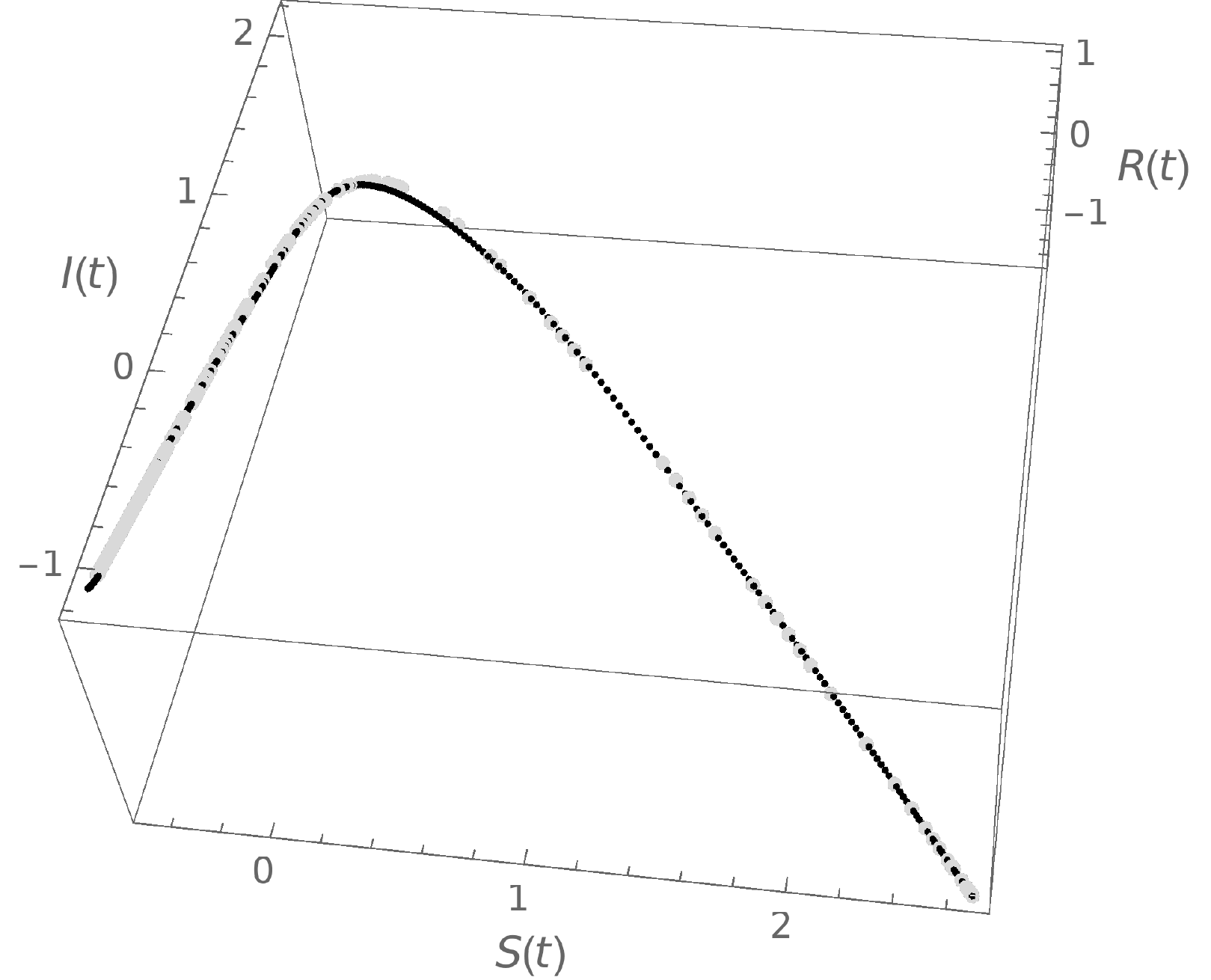}
	\end{center}
	\caption{SIR model. The big lightgry points shows the simulates data and the black pints the orbit generated by the model from the first element of the data. The left figure shows the result using uniformly sampled data and the right figure, the same with nonuniformly sampled data.}
	\label{Fig4}
\end{figure}

Figure \ref{Fig4}. compares the observed and predicted orbit, from the same initial condition and shows the phase protrait generated by the model in the case of this system

\begin{eqnarray}
\dot S(t) &=& -\alpha~ S(t) ~ I(t)  \nonumber \\
\dot I(t) &=& \alpha ~S(t) I(t) -\beta I(t),  \nonumber \\
\dot R(t) &=&  \beta ~I(t) - \gamma~ I(t),
\end{eqnarray}	

\subsubsection{Chua's system}
Chua's system is a ODE system 

\begin{eqnarray}
\dot x(t) &=& \alpha~ (y(t) - x(t) - f(x)) \nonumber \\
\dot y(t) &=& x(t) - y(t) + z(t),  \nonumber \\
\dot z(t) &=&  \beta~ y(t) - \gamma~ z(t),
\end{eqnarray}	

\noindent
with,

\begin{equation}
f(x) = \frac{1}{2} (\left| x+1\right| -\left| x-1\right| ) (m_0-m_1) + m_1 x
\end{equation}

\noindent
representing a simple electronic circuit that exhibits classic chaotic behavior. Here we have choose a set of parameter values which give chaotic solutions: $ \alpha=9.35159085$, $\beta=14.790319805$, $\gamma = 0.016073965$, $m_0 = -1.138411196$, $m_1 = -0.722451121$.

This system presents a strong sensitivity to initial conditions, which makes long-term predictions impossible, as well as structural instability, in the sense that small changes in the parameters lead to totally dissimilar evolution. For these reasons, it constitutes a good benchmark for the validation of models.

\begin{figure}[h]
	\begin{center}
		\includegraphics[width=5cm, height=5cm]{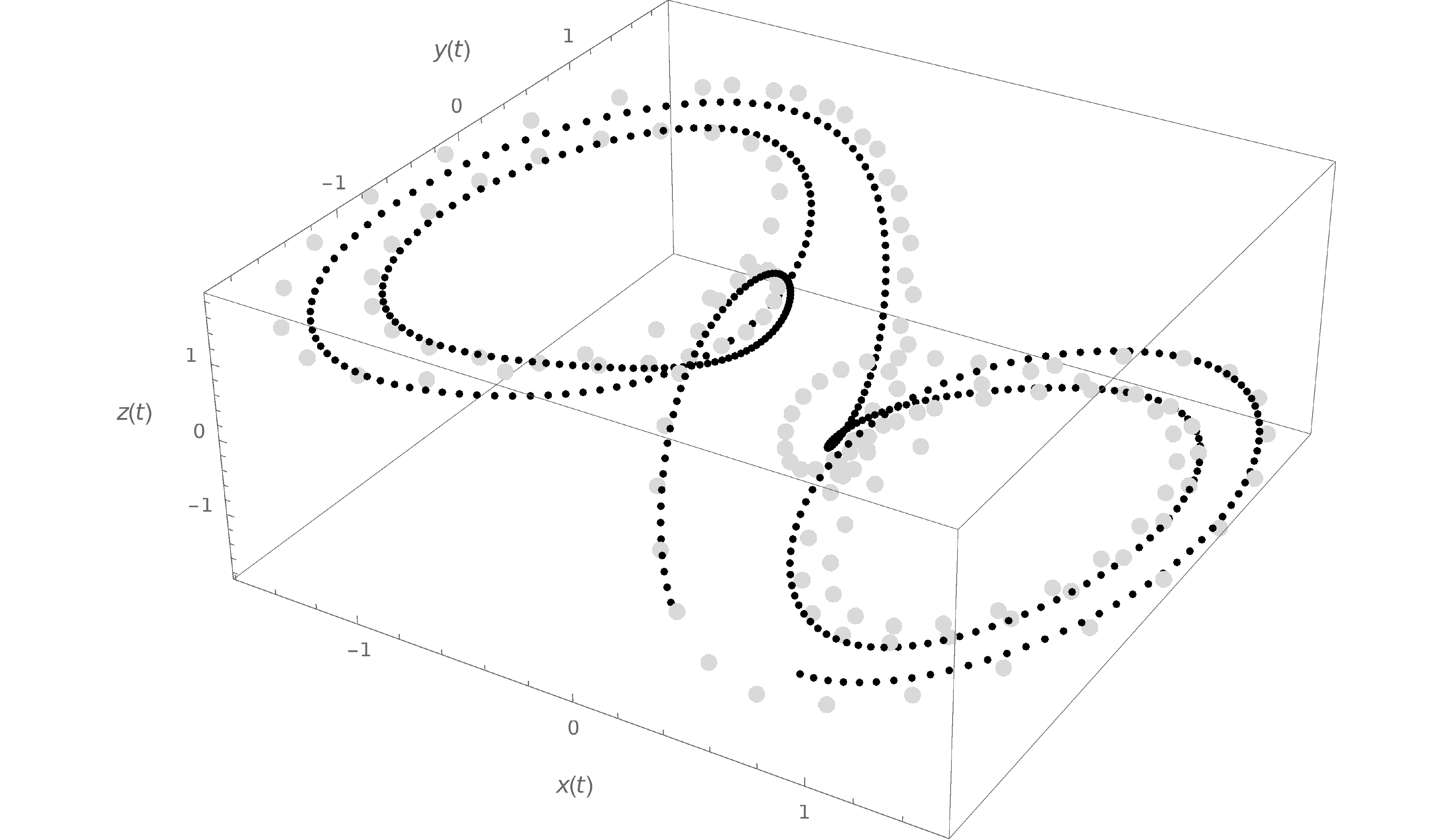}~~
		\includegraphics[width=4.5cm, height=5cm]{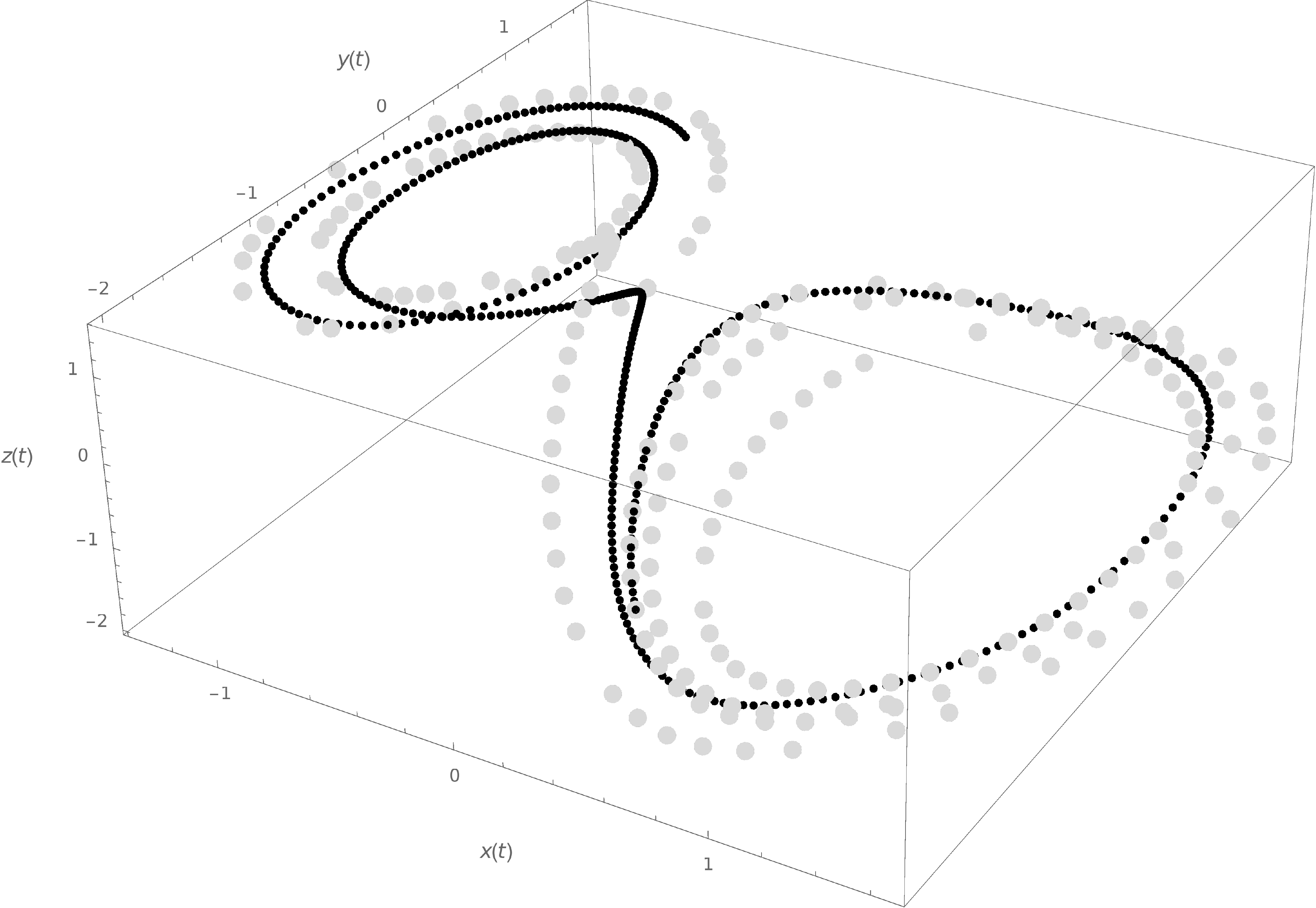}
	\end{center}
	\caption{Chua's system. The big lightgry points shows the simulates data and the black pints the orbit generated by the model from the first element of the data. The left figure shows the result using uniformly sampled data and the right figure, the same with nonuniformly sampled data.}
	\label{Fig5}
\end{figure}	

Figure \ref{Fig5} compares the observed and predicted orbit, from the same initial condition and shows the phase protrait generated by the model in the case of this system.

\subsection{Experimental data}		
To show the performance of this strategy in the case of real (noisy) data, we have chosen two data sets. One, the Hare-Liynx data, is a set whose characterization is, for as many reasons very important and other that stands out for its current relevance, data associated with the COVID-19 epidemic.

\subsubsection{Predator-Prey data}	
This time series is give by the numerical fluctuations in the populations of Canadian lynx (Lynx canadensis) and snowshoe hare (Lepus americanus) caught and then purchased by the Hudson Bay Company in Canada from $1910$ to $1935$ for the American fur market\cite{MacLulich}.

\begin{figure}[h]
	\begin{center}
		\includegraphics[width=6.5cm, height=5cm]{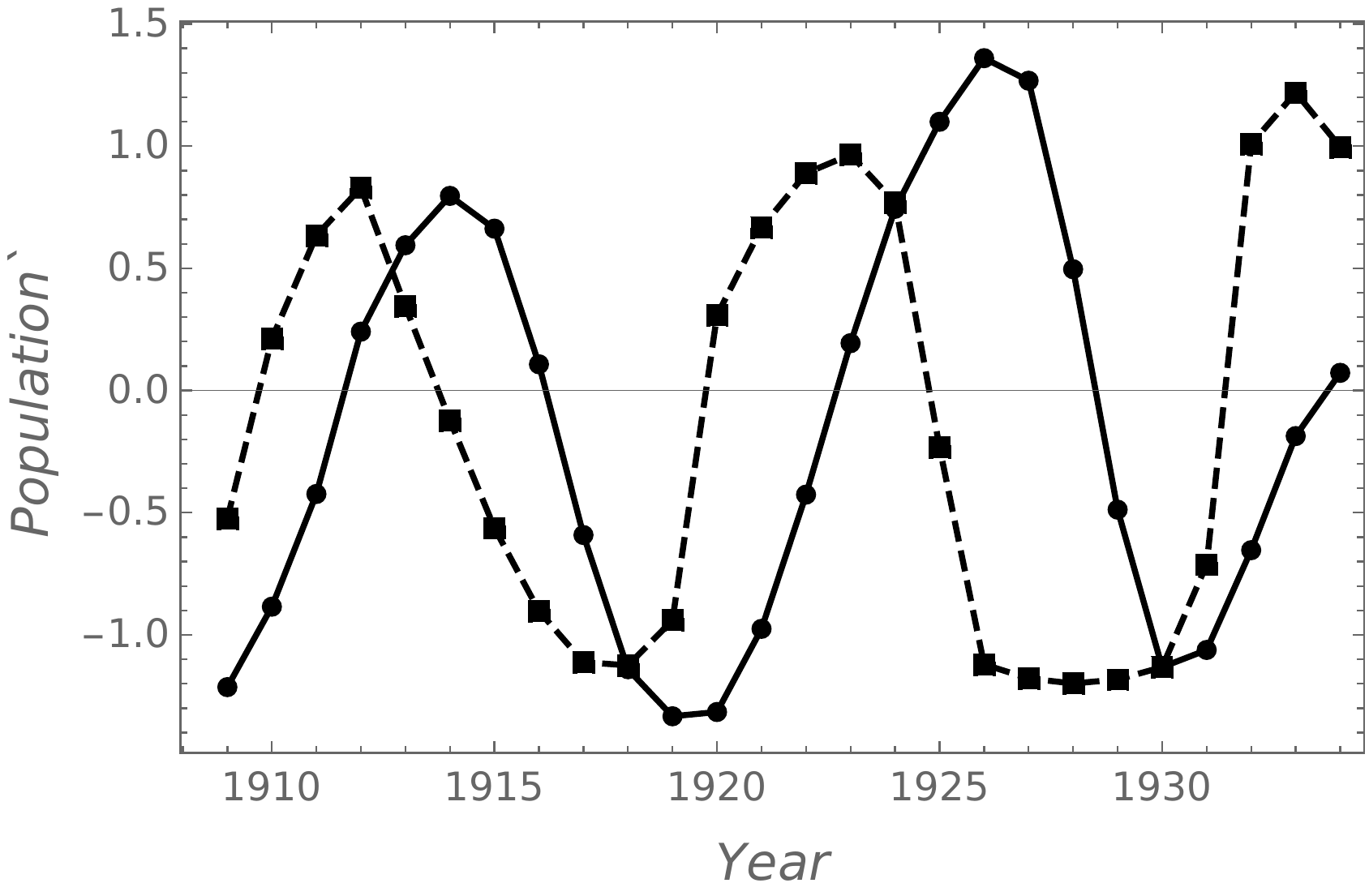}~~
		\includegraphics[width=5.5cm, height=5cm]{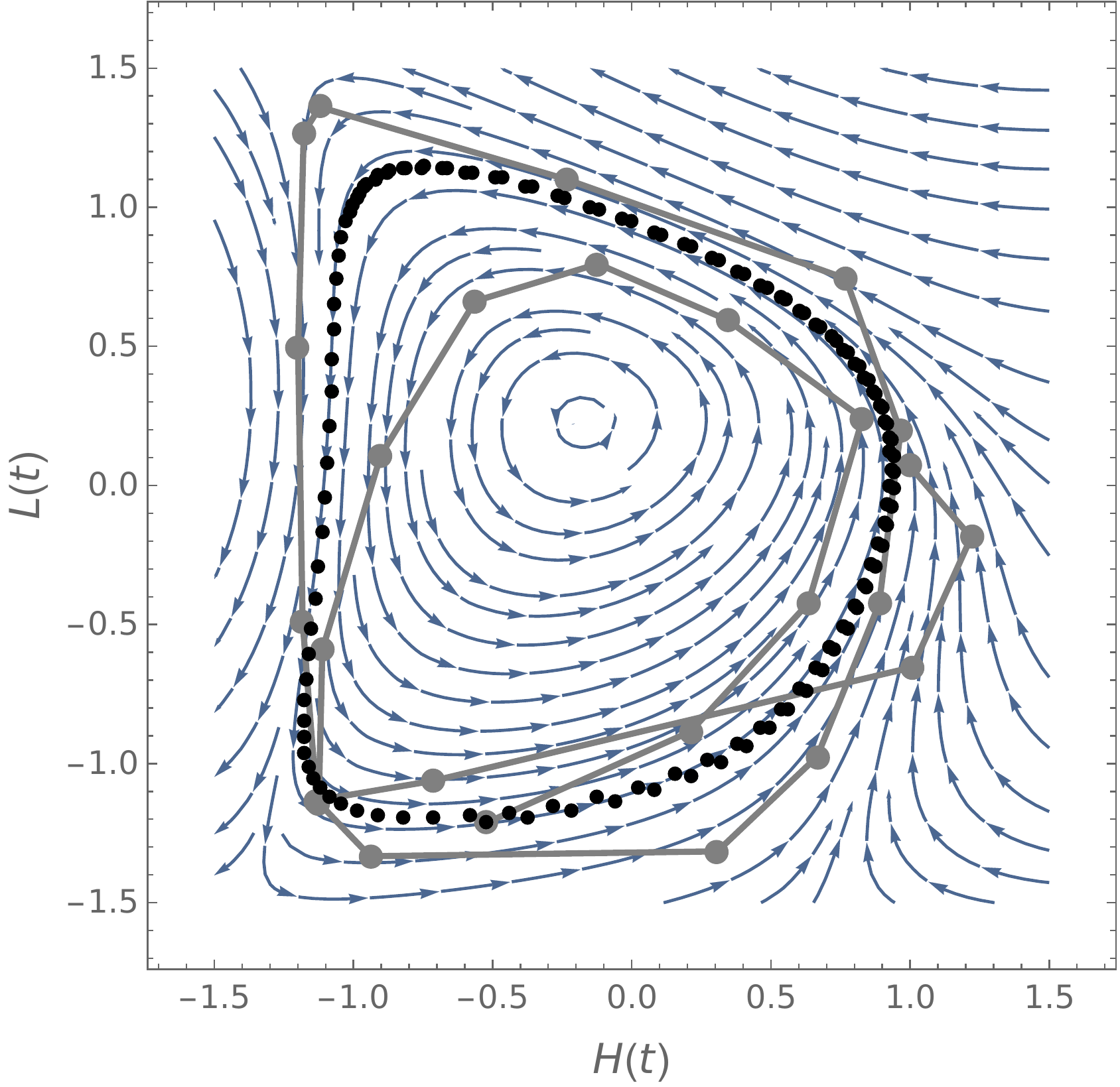}
	\end{center}
	\caption{Predator-Prey model. The figure on the left shows the data and the figure on the right shows the phase portrait generated by the model. In this figure, the large gray dots show the data and the black ones, the trajectory generated by the model from  an initial condition equal to the first element of the data.}
	\label{Fig6}
\end{figure}	

Figure \ref{Fig6} compares the observed and predicted orbit, from the same initial condition and shows the phase protrait generated by the model in the case of this system.

\subsubsection{Coronavirus disease 2019}		
This is data in csv format, updated daily from https://github.com/CSSEGI SandData/COVID-19, maintained by Johns Hopkins University Center for Systems Science and Engineering (CSSE). 		
The data available has information about Confirmed, Recovered and Death people, in our case we transform, the data as: $S = N_p - Confirmed$, $I = Confirmed - Recovered - Deaths$ and $R = Recovered + Deaths$. Here $N_p$ is the population of the country in each case.	

\begin{figure}[h]
	\begin{center}
		\includegraphics[width=4.5cm, height=4.5cm]{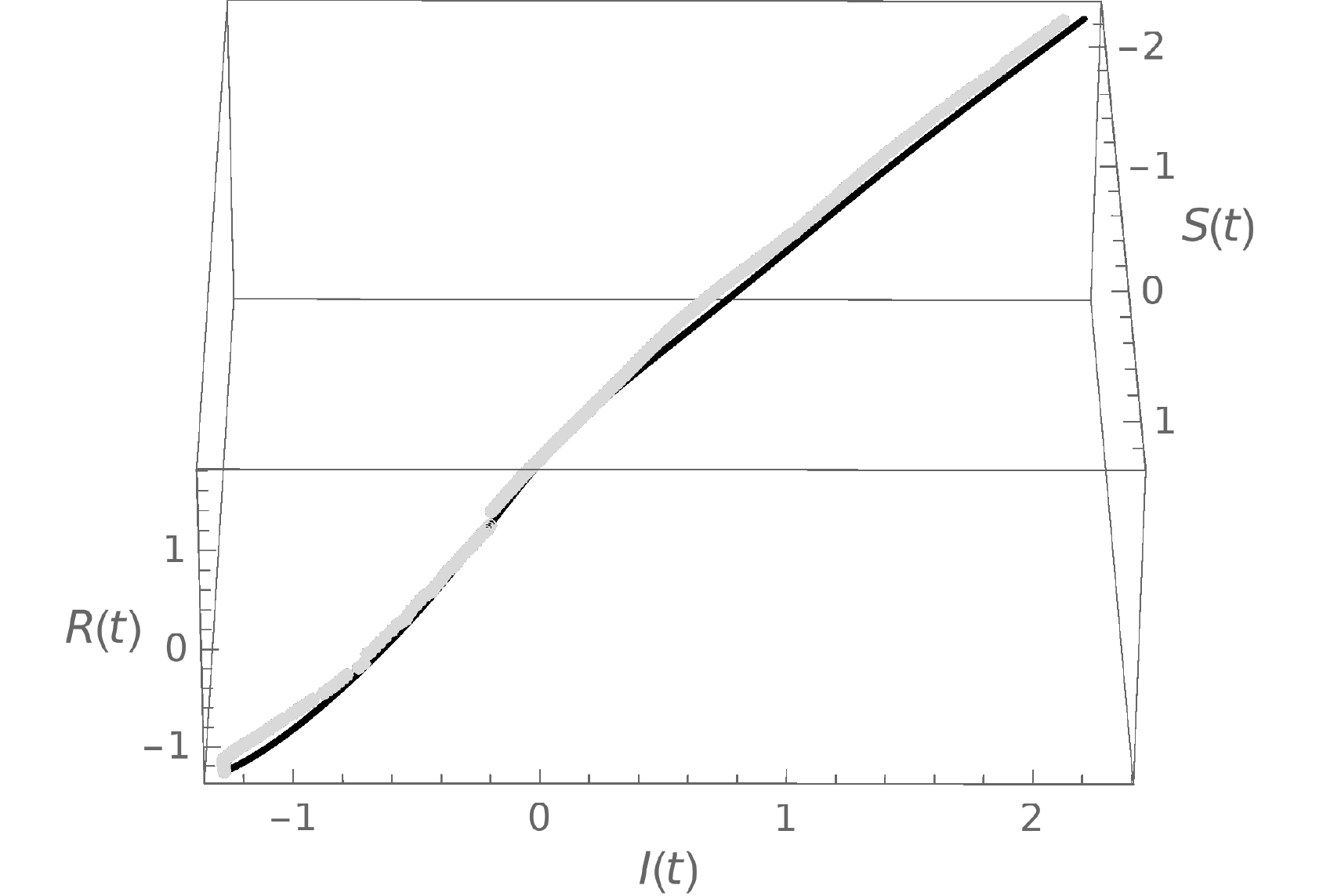}
		\includegraphics[width=4.5cm, height=4.5cm]{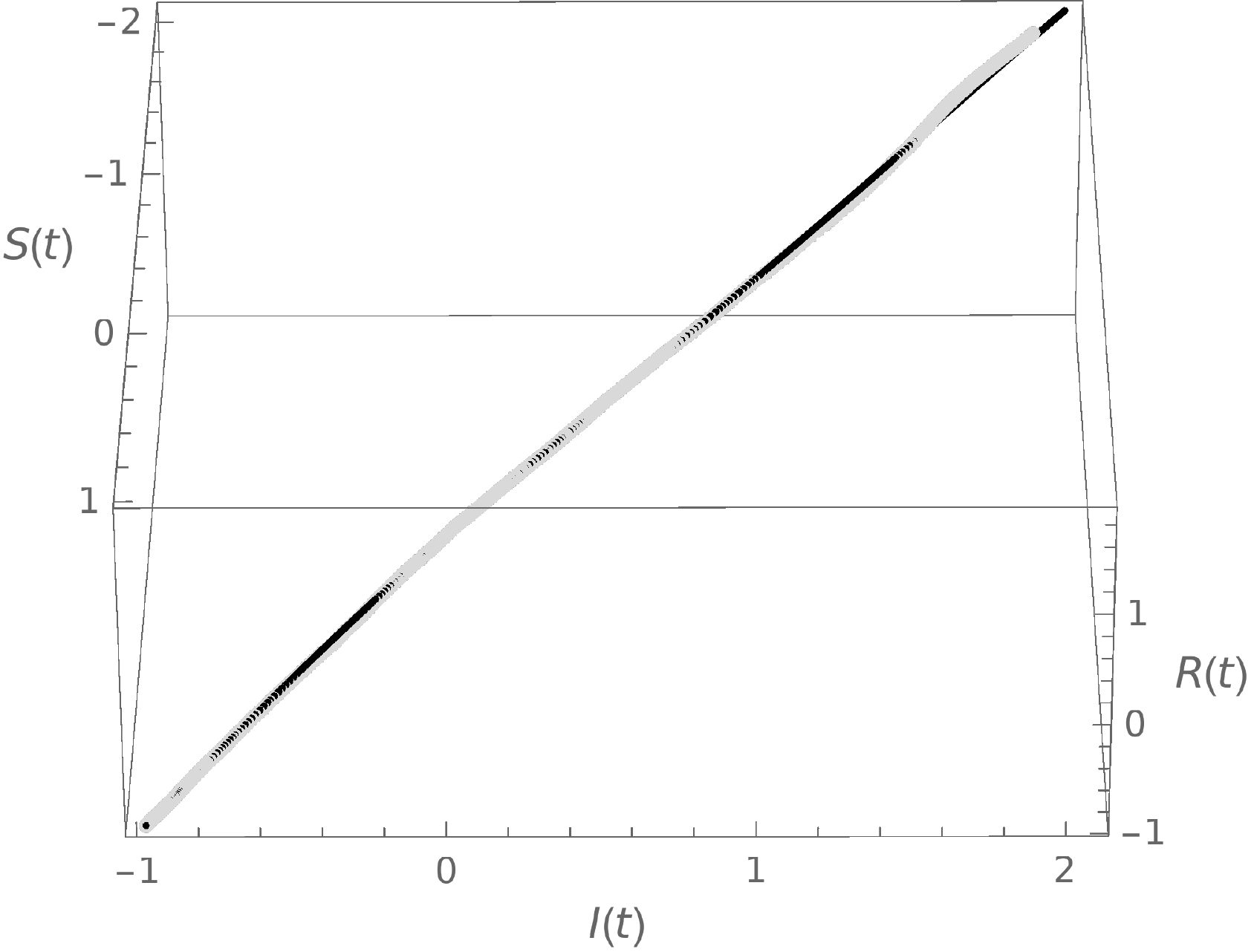}\\
		\includegraphics[width=4.5cm, height=4.5cm]{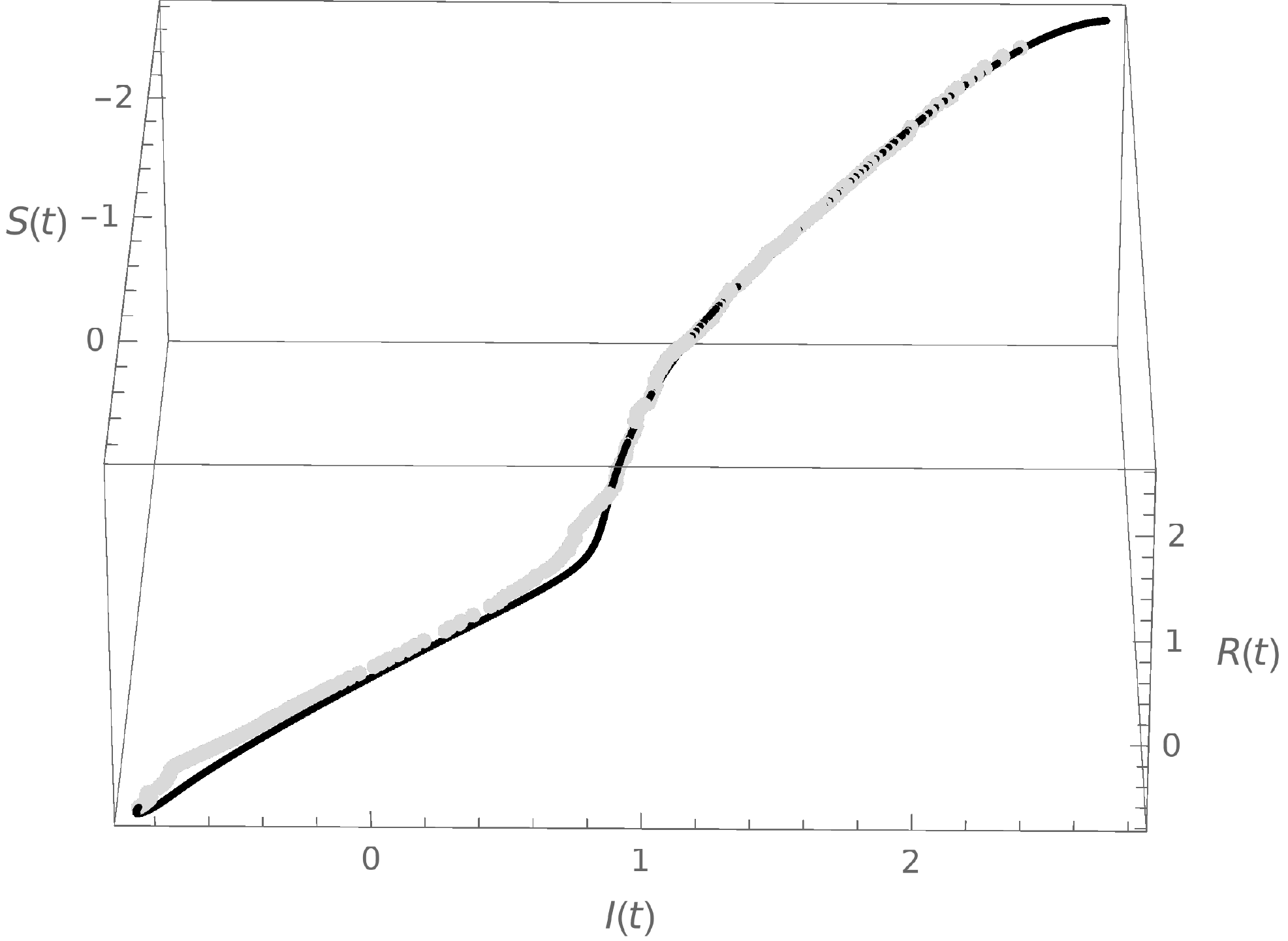}
		\includegraphics[width=4.5cm, height=4.5cm]{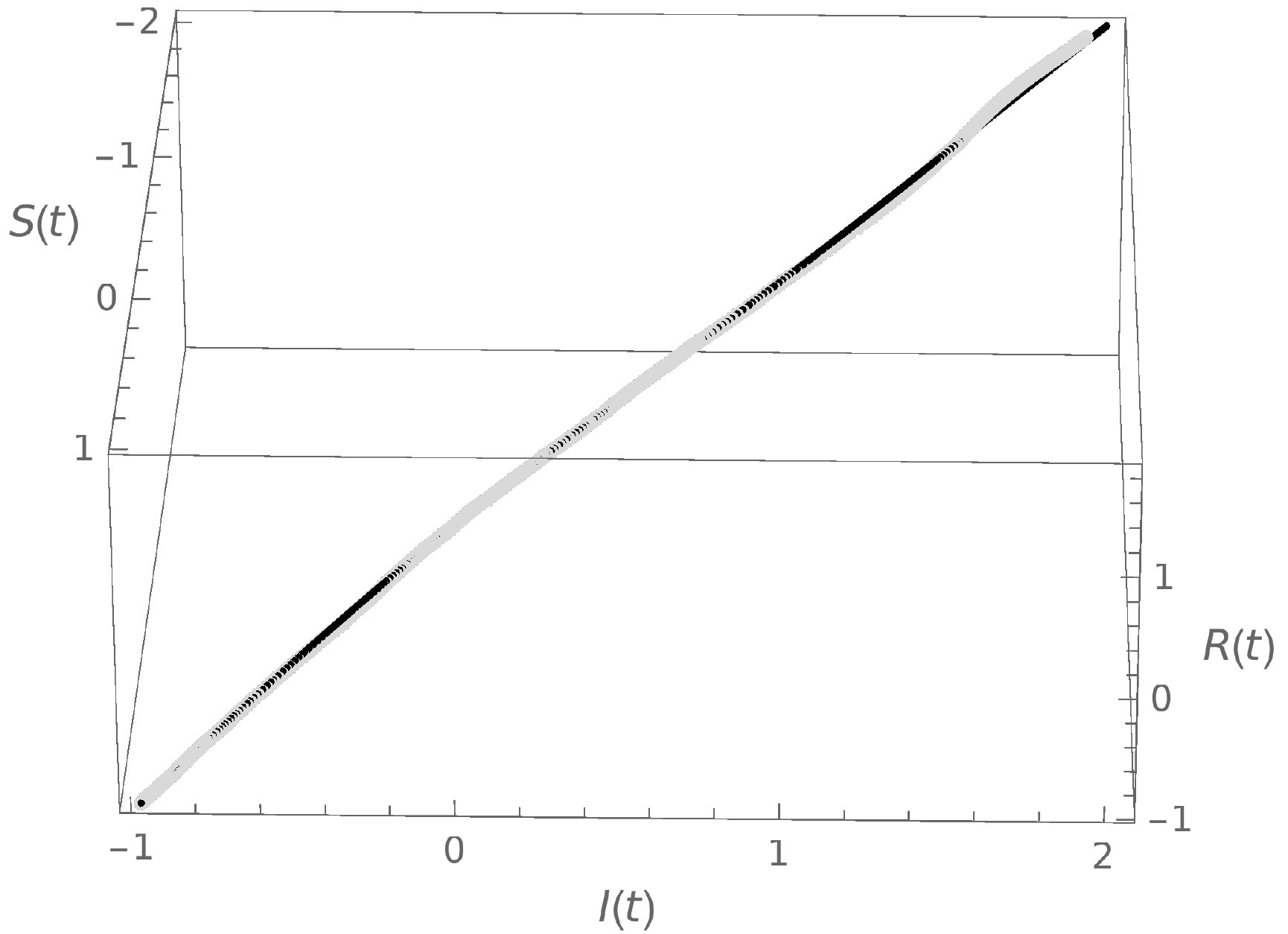}\\
		\includegraphics[width=4.5cm, height=4.5cm]{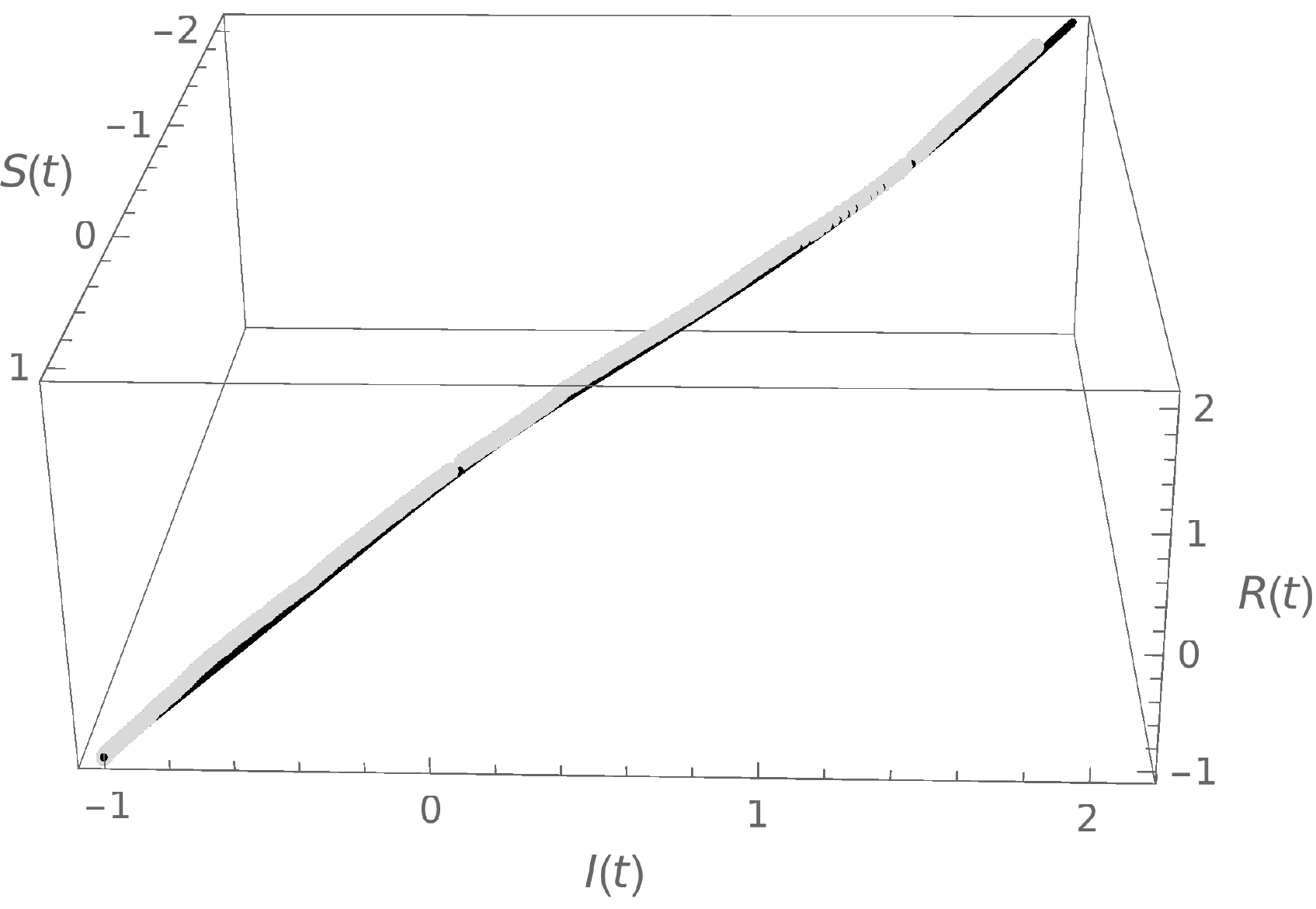}
		\includegraphics[width=4.5cm, height=4.5cm]{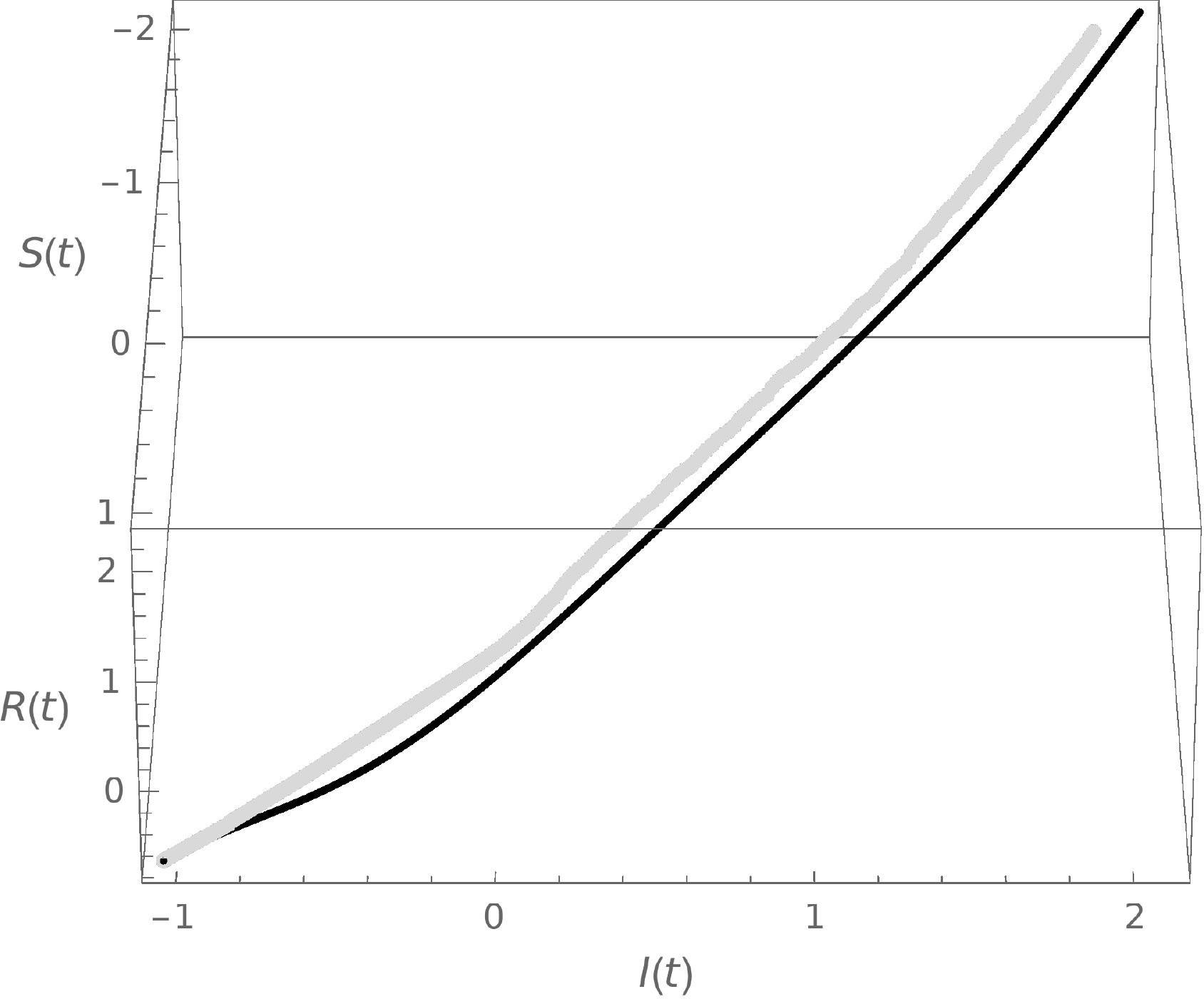}\\
		\includegraphics[width=4.5cm, height=4.5cm]{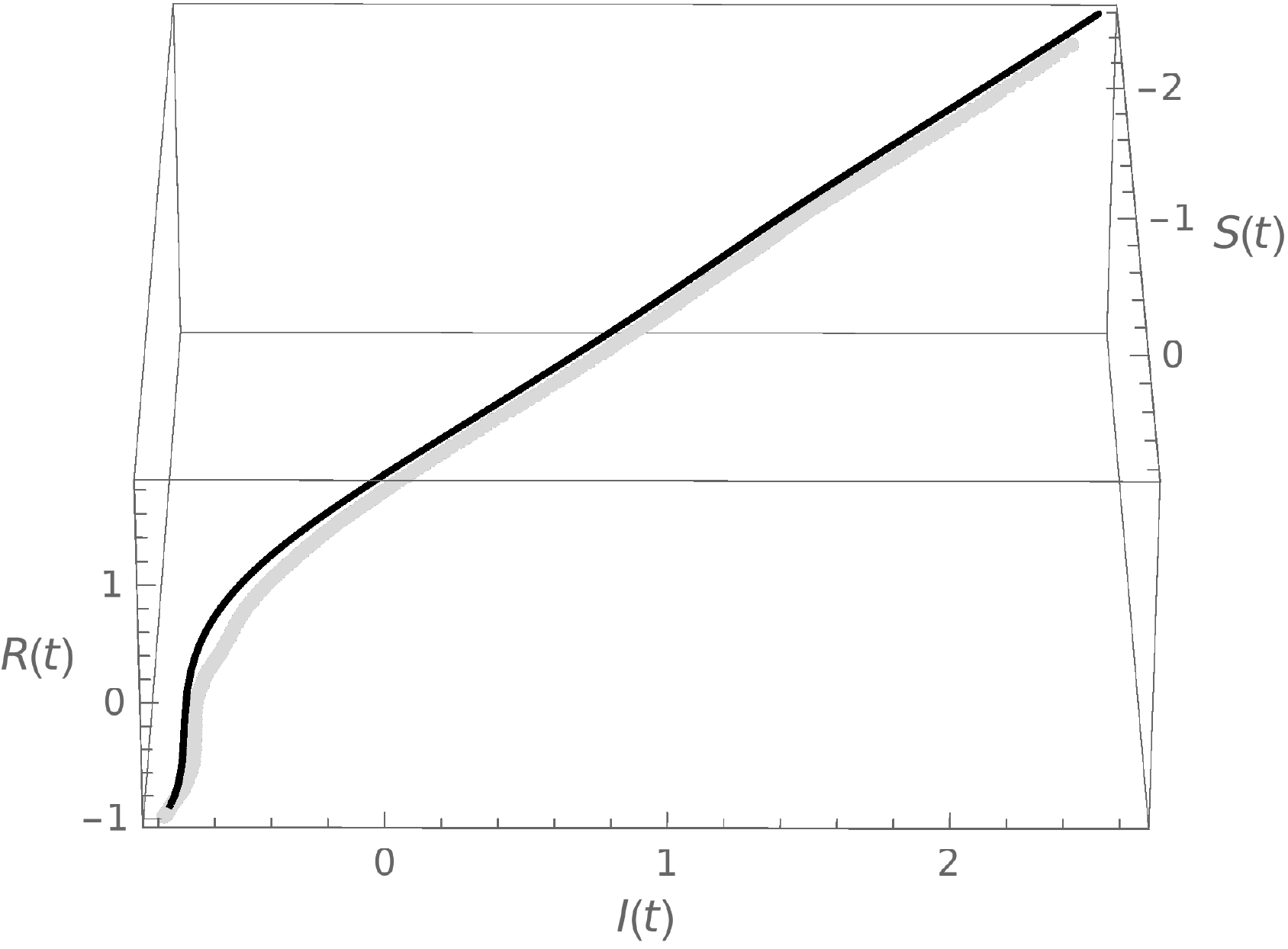}
		\includegraphics[width=5.5cm, height=4.5cm]{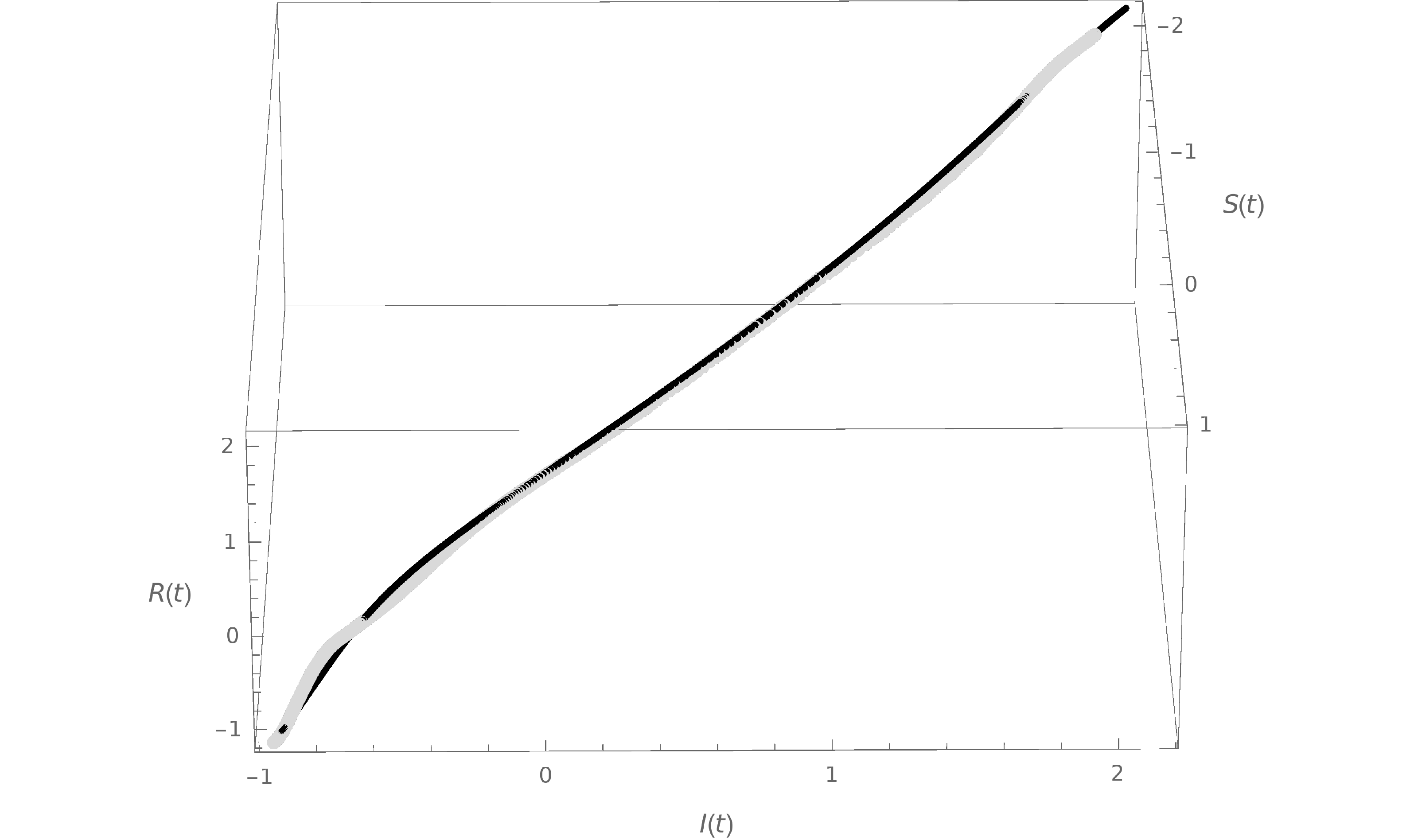}
	\end{center}
	\caption{SIR models from experimental data. The  two colums of figures content the results of the modeling for data from 8 countrys. The first column the results correspond to: Chile,  France, Mexico and Spain  and in the second one for: Colombia, Japan,  Rusia  and USA. The big gray points shows the observed data and the black points, the orbit generated by the model from first element of the data.}
	\label{Fig7}
\end{figure}

Figure \ref{Fig7} compares the observed and predicted orbit, from the same initial condition.

\section{Concluding remarks}
We have presented a strategy to design continuous-time models from  data. This models allows approximate predict the state of the system at any time (whiting a finite interval) and as far as we know, is the first time that a continuous model has been designed using the kernel methods.

To summarize our findings, we will highlight two aspects regarding the performance of proposed strategy:  the idea is simple and the algorithm is too, the dynamics of the system is approximated from data and data may contain noise or be non-uniform sampled.

The strategy allows incorporating any method of integration into any on-line regression scheme of the model parameters, such as Lasso regression o Gaussian process.

Finally, we believe that it is possible to incorporate additional information, besides the data, to improve the performance of the model, like in physics-informed neural networks\cite{Raissi}.

\end{document}